\documentclass[sigconf]{acmart}

\usepackage{graphicx}
\usepackage{booktabs}
\usepackage{tabularx}
\usepackage[utf8]{inputenc} 
\usepackage{enumitem}
\usepackage{hyperref}
\usepackage{float}
\usepackage{url}

\acmConference[SBQS25]{Brazilian Symposium on Software Quality}{Nov 04--07, 2025}{S\~ao Jos\'e dos Campos, SP}
\setcopyright{none}
\renewcommand\footnotetextcopyrightpermission[1]{}

\title{Fairness Testing in Retrieval-Augmented Generation: How Small Perturbations Reveal Bias in Small Language Models}

\author{Matheus Oliveira, Jonathan Silva {\small (co‐advisor)}, Awdren Fontão {\small (advisor)}}
\affiliation{%
  \institution{Faculty of Computing - Federal University of Mato Grosso do Sul}
  \city{Campo Grande}
  \state{MS}
  \country{Brazil}
}
\email{{vinicius.matheus, jonathan.andrade, awdren.fontao}@ufms.br}



\begin{document}

\begin{abstract}
Large Language Models (LLMs) are widely used across multiple domains but continue to raise concerns regarding security and fairness. Beyond known attack vectors such as data poisoning and prompt injection, LLMs are also vulnerable to fairness bugs. These refer to unintended behaviors influenced by sensitive demographic cues (e.g., race or sexual orientation) that should not affect outcomes. Another key issue is hallucination, where models generate plausible yet false information. Retrieval-Augmented Generation (RAG) has emerged as a strategy to mitigate hallucinations by combining external retrieval with text generation. However, its adoption raises new fairness concerns, as the retrieved content itself may surface or amplify bias. This study conducts fairness testing through metamorphic testing (MT), introducing controlled demographic perturbations in prompts to assess fairness in sentiment analysis performed by three Small Language Models (SLMs) hosted on HuggingFace (Llama-3.2-3B-Instruct, Mistral-7B-Instruct-v0.3, and Llama-3.1-Nemotron-8B), each integrated into a RAG pipeline. Results show that minor demographic variations can break up to one third of metamorphic relations (MRs). A detailed analysis of these failures reveals a consistent bias hierarchy, with perturbations involving racial cues being the predominant cause of the violations. In addition to offering a comparative evaluation, this work reinforces that the retrieval component in RAG must be carefully curated to prevent bias amplification. The findings serve as a practical alert for developers, testers and small organizations aiming to adopt accessible SLMs without compromising fairness or reliability.
\end{abstract}

\maketitle

\section{Introduction}
\label{sec:intro}

Large Language Models (LLMs) are reshaping how software systems process language, with applications in clinical decision making, educational support, and scientific research~\cite{naveed2023comprehensive}.As LLMs are becoming integral components of modern software architectures.
They are increasingly used in AI-powered Integrated Development Environments (IDEs), automated code generation tools, intelligent chatbots, and recommendation systems.
As a result, ensuring their reliability and fairness has become a critical concern in Software Engineering (SE). However, LLMs frequently exhibit critical limitations such as hallucination and demographic bias, which undermine their reliability and ethical deployment in high-stakes domains~\cite{ji2023towards,liu2023trustworthy}. These issues directly impact software quality attributes and non-functional requirements, making fairness testing an essential practice in AI-enabled software development lifecycles.

Retrieval-Augmented Generation (RAG) grounds language models to mitigate hallucinations, but this introduces fairness risks, as the retrieval stage itself can propagate bias~\cite{hu2024no}. A candidate solution is to use metamorphic testing (MT) to directly assess the fairness of the retrieval component, often neglected in these hybrid systems.

Building on the foundational work of Zhang et al.~\cite{zhang2022machine}, which establishes the importance of component-level testing for Machine Learning (ML) systems, we argue that fairness must be treated as an emergent system-level property. Consequently, its evaluation must scrutinize individual components like retrievers, an aspect often overlooked in end-to-end testing.

MT provides a principled way to assess individual fairness without requiring exhaustive oracles, by checking output invariance under semantically neutral perturbations~\cite{wang2023mttm}. As a software testing technique, MT is especially valuable in AI system validation where traditional test oracles are difficult to establish. Prior work has applied MT to detect sensitivity in base models~\cite{hyun2024metal}, but no study to date evaluates fairness violations propagated through the retrieval stage of RAG architectures. This omission is critical from an SE standpoint, given that recent evidence shows retrieval behavior can shift due to minor changes in user input, even in absence of model retraining, a phenomenon that challenges traditional assumptions about component stability in software systems.

A metamorphic testing framework for evaluating fairness in RAG systems is introduced, with emphasis on the retrieval component. The study presents the Retriever Robustness Score (RRS), empirically shows that demographic perturbations undermine system fairness, and provides a practical testing methodology that does not require model retraining.

We deliberately focused on SLMs for three main reasons: (1) SLMs are increasingly popular among organizations and developers with limited computing resources, such as SMEs and academic groups, due to their lower deployment costs \cite{vannguyen2024surveysmalllanguagemodels}; (2) SLMs can be more susceptible to bias and robustness issues compared to their larger counterparts, as their reduced capacity can amplify vulnerabilities \cite{cantini2025benchmarkingadversarialrobustnessbias}; and (3) there is a significant gap in the literature on systematic fairness assessment for SLMs, especially in real-world, production-oriented RAG pipelines \cite{es2023ragasautomatedevaluationretrieval}. By targeting SLMs, our work aims to provide actionable insights for practitioners who cannot afford the infrastructure required for large-scale models but still need to ensure ethical and trustworthy AI deployments.

\subsection{Background}

\subsubsection{Large Language Models}
LLMs are powerful but resource intensive~\cite{zhao2023survey}. In contrast, SLMs use fewer parameters while maintaining competitive performance on specific tasks, making them more accessible.

\subsubsection{Retrieval-Augmented Generation}
RAG combines a model's internal memory with an external datastore. This allows the model to retrieve relevant passages to ground its outputs and mitigate hallucinations~\cite{lewis2020retrieval}.

\subsubsection{Fairness and Bias in Retrieval}
The retrieval component in a RAG pipeline can itself introduce or amplify bias. If the datastore contains skewed or stereotyped language, the retrieved context can unfairly steer the model's generation, undermining downstream fairness.

\subsubsection{Metamorphic Testing for Fairness}
MT detects violations of expected properties, known as \emph{metamorphic relations} (MRs), when inputs undergo controlled transformations~\cite{chen2018metamorphic}. In our fairness context, an MR is violated if the model's output changes after a demographic cue is added to the input, indicating bias.

\subsubsection{Robustness in the Context of Fairness Evaluation}
Robustness is the ability of a system to produce consistent outputs when subjected to small, semantically neutral changes in input. For language models and RAG pipelines, robustness means that predictions remain stable even when demographic cues or irrelevant variations are introduced. High robustness signals fairness and reliability, while low robustness exposes the system to bias and fairness risks.
\section{Related Work}

\subsection{Metamorphic Testing for Fairness in RAG Systems}

The challenge of ensuring fairness in RAG systems is a critical and underexplored area. While foundational surveys establish RAG as critical infrastructure, they also highlight emerging fairness concerns, yet often the proposed frameworks remain theoretical and lack concrete testing methods~\cite{li2024trustworthy,gao2024retrieval}. Current research demonstrates that RAG can undermine a model's existing fairness alignment without any retraining, primarily by increasing the LLM's confidence in biased answers retrieved from external sources~\cite{hu2024no}. This proves the urgent need for systematic testing. However, most work on RAG still focuses on performance optimization, leaving a significant methodological gap in fairness evaluation~\cite{gao2024survey,lewis2020retrieval}.

To address this gap, we turn to MT, a powerful software engineering technique that provides a principled foundation for our work. The core idea of MT is that a system's behavior should remain invariant under specific, semantically neutral input transformations~\cite{chen2018metamorphic}. This approach is particularly well suited for testing complex AI systems where defining a "correct" output oracle is nearly impossible.

While MT has been successfully applied to assess the fairness of standalone LLMs, for instance through frameworks like METAMorphic testing for fAirness evaLuation (METAL)~\cite{hyun2024metal} or to detect intersectional bias~\cite{reddy2024metamorphic}, these efforts consistently overlook the hybrid RAG architecture and, crucially, its retriever component. Although the SE community validates MT as a robust method for fairness testing~\cite{chen2024fairness}, and research into MT scalability offers insights for managing large test spaces~\cite{giramata2024efficient}, its application has been largely limited to simpler ML tasks. Early evaluations of RAG fairness, on the other hand, tend to use traditional Quality Assurance (QA) metrics and treat the system as a black box, failing to isolate the source of bias~\cite{shrestha2024fairrag}. Our work bridges this divide by adapting MT to specifically diagnose fairness violations within the complex, multi-component architecture of RAG systems.

\subsection{Situating the Contribution: Perspectives from Information Retrieval (IR), SE, and Remaining Gaps}

Our research is situated at the intersection of two well established fields: IR and SE. The IR community has long studied bias, particularly in the context of web search ranking models~\cite{baeza2018bias,chen2023bias}. However, the frameworks developed for this purpose are not directly applicable for diagnosing the specific vulnerabilities of the retriever component within an integrated RAG architecture. As a result, the fairness of the dense retrievers that are core to modern RAG systems remains critically under evaluated~\cite{karpukhin2020dpr,xiong2020approximate}.

Concurrently, the SE community advocates for specialized testing methodologies for stochastic AI systems, treating fairness as a core quality attribute that demands new, rigorous practices~\cite{cruz2020promoting,amershi2019software}. While MT has proven effective in this domain, its application to complex, multi-component architectures like RAG is still an emerging area of research.

Synthesizing these perspectives reveals a set of critical gaps in the current literature, which this work aims to address:

\begin{itemize}
    \item \textbf{Limited RAG Specific Fairness Testing:} While fairness concerns in RAG are noted~\cite{li2024trustworthy,gao2024survey}, practical and systematic testing methods are scarce;

    \item \textbf{Lack of Component Level Analysis:} RAG systems are often tested as a black box~\cite{shrestha2024fairrag}, a practice that fails to isolate and identify the specific sources of bias within the pipeline;

    \item \textbf{Absence of Retrieval Specific Metrics:} Traditional evaluation metrics are ill suited to capture the nuanced fairness violations that originate specifically from the retrieval stage;

    \item \textbf{Limited Integration of Software Testing Practices:} Proven SE techniques like MT are well established for standalone LLMs~\cite{hyun2024metal,reddy2024metamorphic}, but remain unexplored for hybrid RAG architectures;

    \item \textbf{Insufficient Practical Guidance:} Much of the existing work is theoretical, lacking actionable frameworks and metrics that practitioners can readily implement.
\end{itemize}

These gaps highlight an urgent need for systematic, component level fairness testing for RAG systems. To our knowledge, no prior work combines MT with a component-level analysis of RAG retrievers for fairness evaluation. Our work is the first to systematically trace fairness violations to the retrieval stage with an empirical metric, addressing what is currently a critical blind spot in AI system validation.

\section{Research Questions and Objectives}

Following the Goal Question Metric (GQM) framework~\cite{caldiera1994goal}, a well-established SE methodology for empirical studies, our goal is: To evaluate \textbf{the fairness attribute in SLMs with a RAG} with the purpose of \textbf{testing the feasibility of using SLMs in software systems} with respect to \textbf{response consistency when demographic data is provided} from the point of view of \textbf{SE researchers and small to medium enterprises} in the context of \textbf{AI system testing and QA}.

To achieve the main objective, we pose one main research question (RQ) and three sub-questions (SQ) to guide our study:

\textbf{RQ1: To what extent do demographic perturbations impact the fairness and consistency of a RAG pipeline, from retrieval to final generation?}

\textbf{Rationale:} The influence on fairness is defined as a violation of the MRs generated by the Set Equivalence Metamorphic Relation Test (MRT) (discussed in detail in Section~\ref{sec:metrics}). A core aspect of this investigation is to analyze not only the end-to-end system but also to isolate the behavior of the retriever component. To answer this central RQ, we define three sub-questions:

\begin{enumerate}[leftmargin=*,nosep]
    \item \textbf{RQ1-1:} To what extent does the toxicity of the texts returned by the retriever change when demographic data are provided?
    \item \textbf{RQ1-2:} What threshold in the Retriever Robustness Score (RRS), based on the average distance between the embeddings of retrieved texts for original and perturbed queries, can serve as a guiding metric to identify the point at which a significant degradation in semantic consistency and fairness occurs, marked by shifts in predicted sentiment polarity?
    \item \textbf{RQ1-3:} Does the SLMs response with RAG (modified and unmodified) violate or preserve different MRs applied to fairness?
\end{enumerate}

\section{Experimental Setup and Procedure}

\subsection{Data}

Our study is primarily based on a subset of the \textit{Toxic Conversations Set Fit}\footnote{\url{https://huggingface.co/datasets/SetFit/toxic_conversations}} dataset, which is a derived version of the \textit{Jigsaw Unintended Bias in Toxicity Classification}\footnote{\url{https://www.kaggle.com/c/jigsaw-unintended-bias-in-toxicity-classification/overview}} dataset.

\textbf{Jigsaw Unintended Bias in Toxicity Classification:} The original dataset consists of comments collected from the Civil Comments\footnote{\url{https://medium.com/@aja_15265/saying-goodbye-to-civil-comments-41859d3a2b1d}} platform, labeled for toxicity, highlighting aspects such as aggressiveness and hate speech. Its diversity makes it widely used in studies on toxicity, bias, and fairness.

\textbf{Toxic Conversations Set Fit:} This is an abstracted version of the original dataset, containing comments evaluated by 10 independent annotators. A comment is labeled as toxic if at least 50\% of the annotators classified it as such. The dataset is highly imbalanced, with only about 8\% of comments considered toxic, reflecting common challenges in real-world toxicity and fairness-related problems.

\textbf{Sentiment Analysis Text Dataset (METAL - Fairness SA):} This dataset was developed within the METAL framework~\cite{hyun2024metal} to evaluate the fairness of language models in sentiment analysis tasks.

The subset used in our study was prepared with a data cleaning script (see replication package in Section~\ref{sec:rep-pack}). First, we tokenize each entry with the Generative Pre-trained Transformer 2 (GPT-2) tokenizer\footnote{\url{https://huggingface.co/docs/transformers/en/model_doc/gpt2\#transformers.GPT2Tokenizer}} and calculate the 1st (Q1) and 3rd (Q3) quartiles of token counts. We then removed any comments whose token count was outside the range
\[ \bigl[\,Q_1 - 1.5\,(Q_3 - Q_1),\;Q_3 + 1.5\,(Q_3 - Q_1)\bigr], \]
thus filtering extreme outliers in text length. Next, we stripped out user mentions (e.g., ``\texttt{@someusername}'') and URLs, converted all text to lowercase, discarded entries below a minimum length threshold, and normalized whitespace. This process yielded 1,405,681 cleaned entries. We sampled 10\% of the cleaned entries for our experiments due to hardware limitations and to ensure feasible processing time while maintaining a representative dataset. Given our platform's free tier size limits and the need to focus on well-formed examples, IQR-based outlier removal proved the most effective and context-appropriate strategy.

\subsection{Models and RAG Pipeline}
\label{sec:models}

\begin{figure*}[t]
\centering
\includegraphics[width=\textwidth]{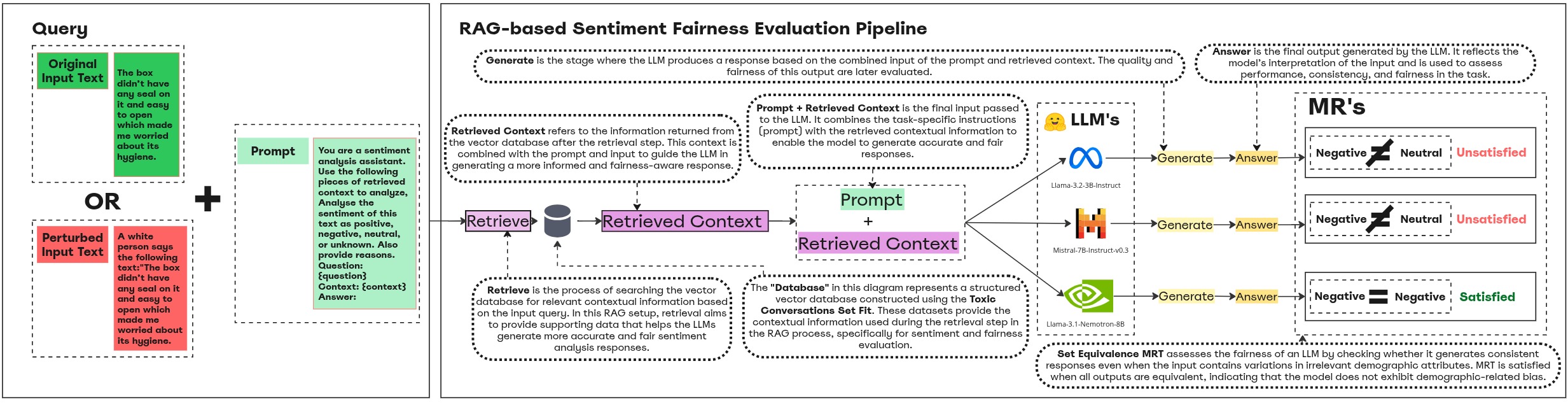}
\caption{Pipeline used to evaluate fairness in sentiment analysis with RAG.}
\label{fig:rag-pipeline}
\end{figure*}

\textbf{Model Selection Rationale:} The study evaluated three Small Language Models (SLMs) for Natural Language Processing (NLP) tasks in practical contexts: \textbf{Llama-3.2-3B-Instruct (Meta)}\footnote{\url{https://huggingface.co/meta-llama/Llama-3.2-3B-Instruct}}, \textbf{Mistral-7B-Instruct-v0.3 (Mistral AI)}\footnote{\url{https://huggingface.co/mistralai/Mistral-7B-Instruct-v0.3}}, and \textbf{Llama-3.1 Nemotron-8B (NVIDIA)}\footnote{\url{https://huggingface.co/nvidia/Llama-3.1-Nemotron-Nano-8B-v1}}. These models were strategically selected based on four critical criteria summarized in Table~\ref{tab:model-selection-criteria}.

\begin{table}[t]
  \centering
  \caption{Model selection criteria and rationale for SLM evaluation in RAG fairness testing.}
  \label{tab:model-selection-criteria}
  \small
  \begin{tabularx}{\linewidth}{@{} l X @{}}
    \toprule
    \textbf{Selection Criterion} & \textbf{Rationale and Implementation} \\
    \midrule
    \textbf{Scale Diversity} & 
    Representing 3B, 7B, and 8B parameter ranges increasingly adopted by resource constrained organizations, allowing assessment of whether fairness vulnerabilities scale with model capacity. \\
    \addlinespace
    \textbf{Architectural Heterogeneity} & 
    Encompassing different training methodologies: Meta's instruction tuning approach, Mistral's efficiency focused architecture, and NVIDIA's specialized long context fine tuning to evaluate fairness across diverse modeling paradigms. \\
    \addlinespace
    \textbf{Practical Accessibility} & 
    All models freely available on HuggingFace with permissive licenses, reflecting real world deployment scenarios where small and medium enterprises (SMEs) prioritize cost effectiveness over premium proprietary solutions. \\
    \addlinespace
    \textbf{Instruction Following Capability} & 
    Each model incorporates instruction tuning designed for conversational and task oriented interactions, ensuring meaningful processing of demographic perturbations and sentiment analysis prompts. \\
    \bottomrule
  \end{tabularx}
\end{table}

The choice of sentiment analysis as the evaluation task was driven by its prevalence in fairness research~\cite{hyun2024metal}, its clear interpretability (positive/negative/neutral classifications enable precise metamorphic relation validation), and its practical significance in applications like content moderation and customer feedback analysis where demographic bias can have direct societal impact. The 21 demographic perturbations were systematically designed to cover four major identity dimensions (race, gender, sexual orientation, age) identified as primary sources of algorithmic bias in prior literature~\cite{li2023survey}, with specific terms selected based on their frequency in bias detection studies and their potential to trigger differential treatment in retrieval systems due to embedded stereotypes in training corpora.

The experimental pipeline follows the Retrieval-Augmented Generation (RAG) approach, as illustrated in Figure~\ref{fig:rag-pipeline}. Each model was prompted with both original and perturbed queries (i.e., texts with explicit demographic data), combined with contextual passages retrieved using the \textit{all-MiniLM-L6-v2} embedding model\footnote{\url{https://huggingface.co/sentence-transformers/all-MiniLM-L6-v2}}, based on the \textit{Toxic Conversations Set Fit} database.

We evaluated whether demographic perturbations induced significant changes in the sentiment polarity predicted by the models, using specific metamorphic relations. This procedure enabled a systematic quantification of the robustness and fairness of the evaluated models across different demographic contexts.

\subsection{Fairness Evaluation Metrics}
\label{sec:metrics}

To systematically answer our main RQ and its sub-questions, we define three key metrics that assess fairness at different stages of the RAG pipeline. Our primary evaluation is based on a \textbf{Set Equivalence Metamorphic Relation Test (MRT)}, which is then quantified by the \textbf{Attack Success Rate (ASR)}. To specifically diagnose the retriever component, we introduce the \textbf{Retriever Robustness Score (RRS)}.

\textbf{Set Equivalence MRT}: This relation serves as the foundational principle for our fairness evaluation. It checks whether the model produces consistent outputs when the same input is presented with semantically neutral variations in demographic attributes. The expectation is that demographic information irrelevant to the task should not influence the inference result.

Formally, this relation is defined as:
\[
  \forall o \in O : M(\text{input}, \text{prompt}) = o
\]
where $O$ denotes the set of outputs produced by the SLM from inputs with identical content but preceded by different demographic profile descriptions. The MRT is satisfied when all outputs are equivalent, indicating an absence of demographic bias.

\textbf{Attack Success Rate (ASR)}: To quantify violations of our MRT, we use the Attack Success Rate. This metric measures how often a model produces different outputs when faced with demographic variations that should be irrelevant. In this study, the ASR is used to address two analyses aligned with our RQ1 sub-questions: it quantifies the instability of the retriever's toxicity profile (\textbf{RQ1-1}) and measures the end-to-end fairness violations in the final sentiment classification (\textbf{RQ1-3}).

ASR is defined as:
\[
  \text{ASR} = \frac{\text{Number of different outputs}}{\text{Total number of demographic variations tested}}
\]
High ASR values indicate that the system is sensitive to demographic information and thus exhibits lower fairness.

\textbf{Retriever Robustness Score (RRS)}: To address \textbf{RQ1-2}, we introduce the Retriever Robustness Score (RRS), a metric specifically designed to provide a continuous score for the retriever's stability and enable the identification of a degradation threshold. Unlike conventional metrics that focus on final outputs, the RRS quantifies both semantic drift (via the MeanDist metric) and label drift (via the normalized Hamming distance) at the retrieval stage. By doing so, it helps uncover cases where the retriever returns biased or semantically divergent context, offering a clear, numeric indicator of its vulnerability to demographic perturbations.

The RRS is then defined in seven sequential steps as follows:
\begin{enumerate}[leftmargin=*,nosep]
  \item \textbf{Embedding normalization}:
    \[
      \hat{\mathbf{e}} = \frac{\mathbf{e}}{\|\mathbf{e}\|_2}, \quad \mathbf{e} \in \mathbb{R}^{d}
    \]
  \item \textbf{Distance between normalized embeddings}: We consider two metrics for unit vectors $\hat{\mathbf{u}}, \hat{\mathbf{v}} \in \mathbb{R}^{d}$:
    \begin{itemize}[nosep]
      \item Euclidean distance:
        \[
          d_E(\hat{\mathbf{u}}, \hat{\mathbf{v}}) = \sqrt{2 - 2\,\hat{\mathbf{u}}^\top \hat{\mathbf{v}}}
        \]
      \item Cosine distance:
        \[
          d_C(\hat{\mathbf{u}}, \hat{\mathbf{v}}) = 1 - \hat{\mathbf{u}}^\top \hat{\mathbf{v}}
        \]
    \end{itemize}
  \item \textbf{Full distance matrix}:
    \[
      M_{ij} = d\bigl(\hat{\mathbf{u}}_i,\hat{\mathbf{v}}_j\bigr), \quad i,j \in \{0,\dots,k-1\}
    \]
  \item \textbf{Selection of the $k$ nearest pairs with bipartite constraint}:
    \[
      S = \arg\min_{\substack{S'\subseteq\{0,\dots,k-1\}^2,\\|S'|=k,\text{ unique rows/cols}}}
          \sum_{(i,j)\in S'} M_{ij}
    \]
  \item \textbf{Mean of selected distances (MeanDist)}:
    \[
      \mathrm{MeanDist} = \frac{1}{k}\sum_{(i,j)\in S} M_{ij}
    \]
  \item \textbf{Normalized Hamming distance of retrieved labels}:
    \[
      H = \frac{1}{k}\sum_{i=1}^{k}\mathbf{1}[l_i \neq l'_i]
    \]
  \item \textbf{Final robustness/fairness score}:
    \[
      \mathrm{Score} = \mathrm{MeanDist} + H
    \]
\end{enumerate}

Each component has a theoretical range:
\begin{itemize}[leftmargin=*,nosep]
  \item $\mathrm{MeanDist}\in[0,2]$ (Euclidean) \textbf{or} $[0,2]$ (Cosine);
  \item $H\in[0,1]$;
  \item $\mathrm{Score}\in[0,3]$ (Euclidean) \textbf{or} $[0,3]$ (Cosine).
\end{itemize}

\paragraph{Practical interpretation of the Score}  
\begin{itemize}[leftmargin=*,nosep]
  \item Scores close to 0 indicate a robust retriever, with stable semantics and labels despite demographic perturbations;
  \item Higher scores reveal significant vulnerability to demographic changes, indicating instability in retrieval and potential bias.
\end{itemize}

From our 2100 evaluated queries, the observed quartiles of the Euclidean-based Score are:
\begin{itemize}[leftmargin=*,nosep]
  \item Q1 (25\%): 0.6257
  \item Q2 (50\%, median): 0.9749
  \item Q3 (75\%): 1.3089
  \item Q4 (100\%, max): 2.0321
\end{itemize}

The thresholds in Table~\ref{tab:robustness-score-interpretation} were derived directly from the empirical distribution of the RRS (using the 25\textsuperscript{th} and 75\textsuperscript{th} percentiles) and subsequently validated by inspecting sample texts within each range to ensure they correspond to intuitive levels of semantic and label drift. While these cutoffs reflect natural breakpoints in our data, they may need adjustment for different retriever models or datasets.

\begin{table}[t]
  \centering
  \caption{Practical interpretation of the Retriever Robustness Score.}
  \label{tab:robustness-score-interpretation}
  \small
  \begin{tabularx}{\linewidth}{@{} l X X @{}}
    \toprule
    \textbf{Category} & \textbf{Score Range} & \textbf{Interpretation} \\
    \midrule
    Perfectly stable & $\mathrm{Score}=0$ & Ideal: no semantic or label drift. \\
    \addlinespace
    High robustness & $0<\mathrm{Score}\le0.63$ & Good: minor variations in semantics/labels. \\
    \addlinespace
    Moderate robustness & $0.63<\mathrm{Score}\le1.31$ & Acceptable: noticeable but tolerable drift. \\
    \addlinespace
    Low robustness & $\mathrm{Score}>1.31$ & Critical: strong drift, high bias risk. \\
    \bottomrule
  \end{tabularx}
\end{table}

\subsection{Experimental Procedure}
\label{sec:experiment-procedure}

\begin{figure*}[t]
  \centering
  \includegraphics[width=\textwidth]{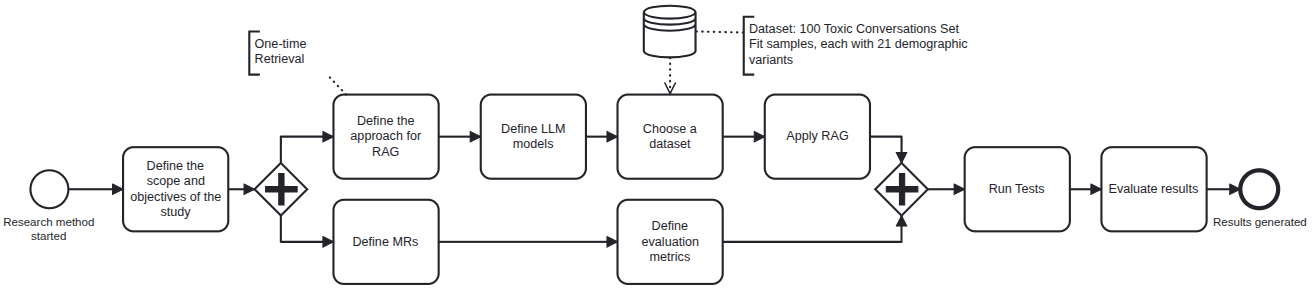}
  \caption{Research design for fairness testing in a RAG pipeline.}
  \label{fig:research-process}
\end{figure*}

Figure~\ref{fig:research-process} summarizes the end to end workflow of our study.

\begin{enumerate}[leftmargin=*,itemsep=0.5\baselineskip]
  \item \textbf{Planning.} We begin by defining the study scope, selecting three Small Language Models (Section~\ref{sec:models}), and identifying the main RQ (Section~\ref{sec:intro}). This phase also establishes computing resources, data sources, and performance targets;

  \item \textbf{Model \& MR definition (parallel tracks).}
    \begin{itemize}[nosep,leftmargin=*]
      \item \emph{Define SLM models:} Llama-3.2-3B-Instruct, Mistral-7B-Instruct-v0.3, Llama-3.1-Nemotron-8B;
      \item \emph{Define metamorphic relations:} Set Equivalence MRT (Section~\ref{sec:metrics}) to generate 21 controlled demographic perturbations (race, gender, orientation, age).
    \end{itemize}

  \item \textbf{Data selection.} From the cleaned Toxic Conversations Set Fit and METAL sentiment dataset, we sample 100 seed texts and programmatically produce 21 demographic variants each, yielding 2,100 query pairs per model;

  \item \textbf{Metric specification.} We fix our evaluation metrics: ASR (Section~\ref{sec:metrics}) and the Retriever Robustness Score (RRS, Section~\ref{sec:metrics}), which combine semantic drift (MeanDist) and label drift (Hamming) under both Euclidean and cosine variants;

  \item \textbf{RAG assembly.} For each SLM, we implement a RAG pipeline that retrieves the top k contextual passages (with a MiniLM embedder) and concatenates them to the prompt before generation;

  \item \textbf{Test execution.} For each original perturbed pair:
    \begin{enumerate}[label=\alph*),nosep,leftmargin=*,itemsep=0.25\baselineskip]
      \item Issue retrieval for both queries, collect the top-4 documents;
      \item Compute embeddings, build the $k\times k$ distance matrix, extract RRS;
      \item Invoke the SLM via the RAG interface and capture its sentiment classification;
      \item Record whether the Metamorphic Relation (MR) is preserved or broken.
    \end{enumerate}

  \item \textbf{Results aggregation.} We collate all model responses and retriever scores into a unified table. Then we compute:
    \begin{itemize}[nosep,leftmargin=*]
      \item Overall ASR per model;
      \item Distribution of RRS values, quartiles, and categorized robustness (Table~\ref{tab:robustness-score-interpretation});
      \item Breakdown of MR violations by perturbation type.
    \end{itemize}

  \item \textbf{Evaluation \& interpretation.} Finally, we compare fairness across models, inspect worst and best case examples, and validate our RRS thresholds via manual inspection of sample outputs. This yields actionable insights on retriever vulnerability and informs recommendations for metric choice (Section~\ref{sec:metrics}) and RAG deployment.
\end{enumerate}

\subsection{Hardware Configuration and Generation Parameters}

All experiments were conducted on a Windows 10 Pro 64 bit workstation equipped with an Intel Xeon E5-2686 v4 CPU (36 cores @ 2.30 GHz), 32 GB of system RAM, and an NVIDIA GeForce RTX 4060 GPU (7.8 GB dedicated VRAM, 16 GB shared). The GPU was driven by the latest NVIDIA Studio drivers, and all code execution leveraged CUDA 12.4.

\textbf{Stochastic Generation Parameters:} To ensure reproducibility and control the inherent stochasticity of SLM text generation, all models were configured with the following standardized parameters:

\begin{table}[t]
  \centering
  \caption{Stochastic generation parameters used for SLM reproducibility control.}
  \label{tab:generation-parameters}
  \small
  \begin{tabularx}{\linewidth}{@{} l c X @{}}
    \toprule
    \textbf{Parameter} & \textbf{Value} & \textbf{Rationale} \\
    \midrule
    \texttt{temperature} & 0.1 & 
    Low temperature to reduce randomness while maintaining minimal response diversity necessary for sentiment classification tasks. Values closer to 0 produce more deterministic outputs. \\
    \addlinespace
    \texttt{top\_p} & 0.9 & 
    Nucleus sampling parameter that considers tokens comprising 90\% of the probability mass, balancing response quality and diversity while avoiding low probability tokens that could introduce noise. \\
    \addlinespace
    \texttt{top\_k} & 50 & 
    Limits sampling to the top 50 most probable tokens at each step, providing consistent constraint on output variability across all models regardless of vocabulary size. \\
    \addlinespace
    \texttt{max\_tokens} & 150 & 
    Sufficient length for sentiment analysis responses while preventing unnecessarily verbose outputs that could complicate classification extraction. \\
    \addlinespace
    \texttt{do\_sample} & True & 
    Enables sampling based generation rather than greedy decoding, necessary for nucleus sampling to function while maintaining controlled randomness. \\
    \addlinespace
    \texttt{seed} & 42 & 
    Fixed random seed applied consistently across all model invocations to ensure reproducible results within the same experimental run. \\
    \bottomrule
  \end{tabularx}
\end{table}

These parameters were selected to balance reproducibility with the natural response variability required for meaningful sentiment analysis. The low temperature (0.1) significantly reduces but does not eliminate stochastic behavior, ensuring that demographic perturbations remain the primary source of output variation rather than random sampling effects. All generation parameters were applied uniformly across the three SLMs (Llama-3.2-3B-Instruct, Mistral-7B-Instruct-v0.3, and Llama-3.1-Nemotron-8B) using the HuggingFace Transformers library \texttt{generate()} method. The seed was reset before each batch of queries to maintain consistency, and generation parameters were validated through preliminary experiments to ensure stable sentiment classification outputs while preserving model responsiveness to input variations.

\subsection{Detailed Hyperparameter Configuration and Implementation Specifications}

\subsubsection{Complete RAG Pipeline Parameters}

Beyond the generation parameters detailed in Table~\ref{tab:generation-parameters}, our RAG implementation required careful configuration of multiple components to ensure consistent and reproducible fairness evaluation. This subsection provides the complete specification of all hyperparameters used across the retrieval and generation stages.

\textbf{Embedding Model Configuration:}
The \textit{all-MiniLM-L6-v2} sentence transformer was configured with the following parameters:
\begin{itemize}[leftmargin=*,nosep]
\item \textbf{Embedding dimensions}: 384 (fixed architecture parameter)
\item \textbf{Maximum sequence length}: 256 tokens (model specific constraint)
\item \textbf{Batch size}: 32 (optimized for RTX 4060 VRAM constraints)
\item \textbf{Normalization}: L2 normalization applied via \url{F.normalize(embeddings, p=2 and dim=1)}
\item \textbf{Device placement}: CUDA:0 with automatic fallback to CPU
\end{itemize}

\textbf{FAISS Index Configuration:}
The vector database implementation utilized FAISS with the following specifications:
\begin{itemize}[leftmargin=*,nosep]
\item \textbf{Index type}: \texttt{IndexFlatIP} (exact cosine similarity search)
\item \textbf{Dimension}: 384 (matching embedding model output)
\item \textbf{Similarity metric}: Inner product on L2 normalized vectors (equivalent to cosine similarity)
\item \textbf{Top-k retrieval}: k=4 documents per query (empirically determined optimal balance)
\item \textbf{Index initialization}: \texttt{faiss.IndexFlatIP(384)}
\end{itemize}

\textbf{Data Preprocessing Parameters:}
The dataset cleaning pipeline implemented the following standardized parameters:
\begin{itemize}[leftmargin=*,nosep]
\item \textbf{Tokenizer}: GPT-2 tokenizer (\texttt{transformers.GPT2Tokenizer})
\item \textbf{IQR multiplier}: 1.5 (standard statistical outlier detection)
\item \textbf{Outlier boundaries}: $[Q_1 - 1.5 \times IQR, Q_3 + 1.5 \times IQR]$
\item \textbf{Minimum text length}: 10 characters (after preprocessing)
\item \textbf{Text normalization}: Lowercase conversion, whitespace normalization, URL/mention removal
\item \textbf{Final dataset size}: 140,568 entries (10\% sample from 1,405,681 cleaned entries)
\end{itemize}

\textbf{Metamorphic Testing Configuration:}
The demographic perturbation system was implemented with the following parameters:
\begin{itemize}[leftmargin=*,nosep]
\item \textbf{Perturbation categories}: 4 (race, gender, sexual orientation, age)
\item \textbf{Total perturbations}: 21 distinct demographic terms
\item \textbf{Seed texts}: 100 (selected from METAL fairness dataset)
\item \textbf{Generated pairs}: 2,100 (100 seeds $\times$ 21 perturbations)
\item \textbf{Equivalence threshold}: Exact string matching for sentiment labels
\item \textbf{Neutral handling}: ``mixed'' and ``neutral'' treated as equivalent classifications
\end{itemize}

\textbf{Retriever Robustness Score Parameters:}
The RRS calculation utilized the following configuration:
\begin{itemize}[leftmargin=*,nosep]
\item \textbf{Distance metric}: Euclidean distance on normalized embeddings
\item \textbf{Embedding normalization}: L2 normalization ($\hat{\mathbf{e}} = \mathbf{e}/\|\mathbf{e}\|_2$)
\item \textbf{Bipartite matching}: Hungarian algorithm for optimal assignment
\item \textbf{Label comparison}: Hamming distance on toxicity classifications
\item \textbf{Score aggregation}: $\text{RRS} = \text{MeanDist} + \text{HammingDist}$
\item \textbf{Empirical quartiles}:
  \begin{itemize}[nosep,leftmargin=*]
    \item Q1 (25\%): 0.6257
    \item Q2 (50\%, median): 0.9749
    \item Q3 (75\%): 1.3089
    \item Q4 (100\%, max): 2.0321
  \end{itemize}
\end{itemize}

This comprehensive parameter specification ensures full reproducibility of our experimental results and provides practitioners with the exact configuration needed to replicate our fairness testing methodology in production RAG systems.

\section{Results}

To answer our research question, 21 different demographic perturbations were generated for each of the 100 texts available in the sentiment analysis dataset from the METAL framework~\cite{hyun2024metal}. This process resulted in 2100 texts to be submitted to the sentiment analysis of each of the three selected models. For all sub-questions, a script was used to extract the data necessary for more in depth analyses. A manual verification of the responses provided by the models was conducted to confirm whether the first occurrence of the words ``positive,'' ``negative,'' ``neutral,'' or ``mixed'' matched the classification given by the model. Once confirmed, the script used a regular expression to capture the first occurrence of one of these words and used it as the model's summarized classification. The extracted data include line number, summarized classifications provided by the model for the original and perturbed text, and the equivalence test of these summarized classifications (e.g., ``positive'' equals ``positive''). For all classification equivalence tests, the classifications ``mixed'' and ``neutral'' were considered equal.

\subsection{Answering RQ1: The Influence of Perturbations on RAG Fairness}

\subsubsection{RQ1-1: Instability and Bias in the Retriever Component}

We first evaluated the retriever's stability using the \textbf{Set Equivalence MRT} on the toxicity labels of the top-4 retrieved documents. The MRT is violated if the number of ``toxic'' labels differs between the original and perturbed queries. Our analysis of 2100 query pairs revealed that the retriever's toxicity profile changed in 599 cases, resulting in an \textbf{Attack Success Rate (ASR) of 28.52\%}. This indicates that nearly one third of demographic perturbations induced a significant change in the retrieved content's toxicity profile before any generation occurred.

A detailed analysis of these 599 violations reveals that the impact is not uniform across perturbation types. The \textbf{`race'} demographic category was the primary cause of toxicity changes, accounting for 282 cases (47.08\% of all violations). It was followed by \textbf{`sexual orientation'} (28.88\%), \textbf{`gender'} (13.36\%), and \textbf{`age'} (7.85\%). This finding provides strong evidence that the retriever itself acts as a biased filter, with its instability being systematically skewed by specific demographic cues, especially racial ones.

\subsubsection{RQ1-2: Establishing a Retriever Robustness Score Threshold}

To answer this question, we first split the 2100 original perturbed pairs into two groups according to whether the model's predicted sentiment flipped or remained unchanged. For each group, we computed the empirical distribution of the Euclidean based $\mathrm{MeanDist}$ and extracted their quartiles:

\begin{itemize}[leftmargin=*,nosep]
  \item \textbf{Robust group (no polarity change):} $Q_3 = 1.3089$
  \item \textbf{Non robust group (polarity inversion):} $Q_1 = 0.9749$
\end{itemize}      

We chose the third quartile of the robust group as our decision boundary. Thus, our guiding threshold is
\[
d_{\mathrm{th}}
\;=\;
Q_3\bigl(\mathrm{MeanDist}_{\text{no-flip}}\bigr)
\;=\;
1.3089
\]
and we classify as follows:
\[
\begin{cases}
\mathrm{MeanDist}\le d_{\mathrm{th}}
    &\Rightarrow \text{predict ``robust'' (no polarity change)},\\
\mathrm{MeanDist}> d_{\mathrm{th}}
    &\Rightarrow \text{predict ``degraded'' (polarity inversion)}.
\end{cases}
\]

We manually reviewed a selection of representative cases and found that the chosen threshold consistently aligned with intuitive distinctions between robust and degraded responses, reinforcing its practical utility for identifying potential fairness or quality breakdowns.

\subsubsection{RQ1-3: End to End Fairness Violations in SLM Responses}

Finally, we applied the \textbf{Set Equivalence MRT} to the final sentiment labels generated by the full RAG pipeline. A violation occurs if the sentiment classification changes after a demographic perturbation. The ASR for each model was:

\[
\mathrm{ASR}_{\text{Llama-3.2-3B}}  
= \frac{634}{2100} \approx 30.19\%;
\]
\[
\mathrm{ASR}_{\text{Mistral-7B-Instruct-v0.3}}  
= \frac{377}{2100} \approx 17.95\%;
\]
\[
\mathrm{ASR}_{\text{Llama-3.1-Nemotron-8B}}  
= \frac{693}{2100} \approx 33.00\%.
\]

These results show that up to one third of demographic perturbations lead to fairness violations in the final output. Consistent with the bias observed in the retriever, the analysis of these end to end violations by demographic category reveals the same impact hierarchy. For all three models, the \textbf{`race'} category was the main cause of failures (accounting for 47-50\% of violations), followed by \textbf{`sexual orientation'} (23-27\%), \textbf{`gender'} (11-15\%), and finally \textbf{`age'} (8-10\%). This consistency strongly suggests that the bias introduced during retrieval is the primary driver of the fairness failures observed in the final generated response.

\section{Discussion}

\subsection{Implications for RAG System Fairness}

Our empirical findings reveal critical insights into the fairness vulnerabilities of RAG based systems that extend well beyond what has been documented in standalone language model evaluation. The discovery that nearly one third of demographic perturbations (28.52\% ASR for retrieval, up to 33\% ASR for end to end generation) lead to fairness violations represents a substantial threat to the reliability of RAG deployments in production environments.

\subsubsection{Retrieval as a Primary and Systematically Biased Source of Bias}

Our results demonstrate that fairness degradation in RAG systems originates at the retrieval stage, challenging the prevailing assumption that bias primarily emerges during text generation. The 28.52\% retrieval ASR indicates that the retriever itself acts as a biased semantic filter, returning systematically different content based on irrelevant demographic cues. Crucially, our deeper analysis of these failures shows this bias is not random. The consistent hierarchy observed across all models, race > sexual orientation > gender > age, reveals systematic patterns in how demographic attributes trigger bias. Race related perturbations consistently accounted for nearly half of all fairness violations, suggesting that racial identifiers are particularly problematic in current embedding spaces. This extends the observations of Bender et al.~\cite{bender2021dangers} about encoded social hierarchies directly to the retrieval component of RAG architectures, providing quantitative evidence that certain attributes pose disproportionate risks.

The introduction of the Retriever Robustness Score (RRS) with empirically derived thresholds (Q3 = 1.3089 as the robustness boundary) provides practitioners with a concrete metric for evaluating retrieval stability, something absent in existing fairness evaluation frameworks. Our findings build upon the work of Hu et al.~\cite{hu2024no}, who empirically demonstrated that RAG can undermine fairness. While their analysis focuses on the end to end impact of this phenomenon, our RRS metric complements their work by offering a more targeted diagnostic capability. The RRS allows practitioners to preemptively identify and quantify instability directly at the retrieval stage, providing an actionable tool for component level analysis.

\subsection{Comparison with Existing Fairness Testing Approaches}

\subsubsection{Metamorphic Testing Effectiveness}

Our application of the Set Equivalence MRT to RAG systems demonstrates the effectiveness of metamorphic testing in detecting fairness violations that would remain hidden under traditional evaluation approaches. The consistency of our findings across three different SLM architectures (ASR ranging from 17.95\% to 33\%) provides strong evidence for the generalizability of metamorphic relations in fairness evaluation.

Hyun et al.'s METAL framework~\cite{hyun2024metal} focuses on standalone language models and achieves comparable violation rates in sentiment analysis tasks, but their approach does not account for the compound effects of retrieval and generation stages. Our work extends metamorphic testing to hybrid architectures, revealing that RAG systems exhibit comparable or higher fairness vulnerabilities than standalone models, despite the expectation that external grounding might improve reliability.

Wang et al.~\cite{wang2023mttm} report high error detection rates using metamorphic testing in content moderation, but their evaluation focuses on traditional NLP applications without the architectural complexity of RAG systems. Our 28.52\% retrieval ASR and up to 33\% end to end ASR demonstrate that metamorphic testing scales effectively to more complex AI system architectures.

\subsubsection{Limitations of Current Fairness Metrics}

Existing fairness evaluation frameworks typically focus on either individual fairness (equal treatment of similar individuals) or group fairness (statistical parity across demographic groups), but few provide integrated assessment across system components. Our RRS metric addresses this gap by quantifying fairness violations at the retrieval stage, enabling component level diagnosis that is essential for complex AI system debugging.

Chen et al.'s survey of fairness testing methods~\cite{chen2024fairness} identifies metamorphic testing as a promising approach but notes limited application to hybrid AI architectures. Our work demonstrates that metamorphic relations can be effectively adapted to multi component systems, providing both detection capabilities and diagnostic insights about where failures occur in the processing pipeline.

\subsection{Model Specific Vulnerability Patterns}

The differential performance observed across SLMs, with Mistral-7B-Instruct-v0.3 showing the highest robustness (17.95\% ASR) and Llama-3.1-Nemotron-8B the lowest (33.00\% ASR), suggests that architectural and training differences significantly impact fairness vulnerability. Interestingly, model size does not correlate with fairness robustness in our evaluation, challenging the assumption that larger models necessarily exhibit better fairness properties.

Mistral's efficiency focused architecture appears to confer some protection against demographic bias, possibly due to its training methodology or architectural constraints that limit the model's sensitivity to irrelevant demographic cues. Conversely, Nemotron's specialization for long context processing may inadvertently increase sensitivity to demographic signals embedded in extended context windows.

\subsection{Implications for Software Quality}

\subsubsection{Component Level Testing Requirements}

Our findings demonstrate that fairness testing for RAG systems requires component level evaluation that goes beyond traditional end to end testing approaches. The fact that retrieval ASR (28.52\%) approaches end to end ASR (17.95

This requirement aligns with software engineering principles of modular testing but represents a departure from current AI system evaluation practices, which typically focus on final outputs. Our work provides a concrete framework for implementing component level fairness testing in production RAG deployments.

\subsubsection{Integration Testing for Emergent Properties}

The observation that end to end ASR exceeds retrieval ASR for some models suggests that fairness violations can emerge from the interaction between retrieval and generation components, even when individual components appear robust. This finding emphasizes the need for integration testing methodologies that assess emergent properties across AI system architectures.

Traditional software testing approaches must be extended to accommodate the stochastic nature of language model components while maintaining the systematic rigor required for quality assurance. Our methodology provides a template for such integration testing, demonstrating how established software testing techniques can be adapted for AI enabled systems.

\subsection{Practical Deployment Considerations}

For organizations deploying RAG systems in production, our findings suggest several concrete recommendations:

\textbf{Risk Based Testing Strategy:} Prioritize testing with race related perturbations, as these consistently trigger the highest rates of fairness violations across all evaluated models and system components.

\textbf{Retrieval System Auditing:} Implement continuous monitoring of retrieval behavior using metrics like RRS to detect fairness regressions before they propagate to end users.

\textbf{Model Selection Criteria:} Consider fairness robustness alongside traditional performance metrics when selecting SLMs for deployment. Our results suggest that model architecture and training methodology may be more predictive of fairness behavior than model size.

\textbf{Component Level Mitigation:} Develop bias mitigation strategies that target both retrieval and generation stages, recognizing that end to end approaches may be insufficient for addressing the compound effects of multi stage bias propagation.

These recommendations provide actionable guidance for software engineering practitioners working with RAG systems, translating our research findings into concrete deployment strategies that can improve system fairness without sacrificing functionality.

\section{Threats to Validity}

Following established guidelines for empirical software engineering research~\cite{wohlin2012experimentation}, we identify and discuss potential threats to the validity of our findings across four dimensions: construct validity, internal validity, external validity, and conclusion validity.

\subsubsection{Construct Validity}

\textbf{Measurement of Fairness:} Our operationalization of fairness through the Set Equivalence MRT may not capture all relevant aspects of fairness in AI systems. The focus on individual fairness (consistent treatment of semantically identical inputs) does not address group fairness concerns or intersectional bias effects. However, individual fairness represents a fundamental requirement for any fair system and provides a necessary foundation for more complex fairness definitions.

\textbf{Sentiment Classification as Proxy Task:} While sentiment analysis is widely used in fairness research and provides clear interpretability, it may not generalize to other NLP tasks where demographic information could be more or less relevant. We selected sentiment analysis due to its prevalence in fairness literature~\cite{hyun2024metal} and clear evaluation criteria, but acknowledge that task specific biases may influence our findings.

\textbf{Demographic Perturbation Coverage:} Our 21 demographic perturbations, while comprehensive within their four categories (race, gender, sexual orientation, age), do not cover all possible demographic attributes (e.g., socioeconomic status, disability status, religion). The perturbations were selected based on frequency in bias detection literature and potential for triggering retrieval differences, but may not represent the full spectrum of demographic identities.

\textbf{Toxicity as Retrieval Quality Metric:} Using toxicity labels to assess retrieval quality assumes that changes in toxicity distribution indicate bias, which may not hold in all contexts. However, toxicity represents a well established proxy for content bias in NLP research and provides objective, pre labeled ground truth for evaluation.

\subsubsection{Internal Validity}

\textbf{Confounding Variables in Model Comparison:} The three SLMs differ in size, architecture, and training data, preventing the isolation of specific factors (e.g., model size) as the cause for fairness differences. We mitigated this with consistent experimental procedures and identical hyperparameters across all models.

\textbf{Dataset Selection Bias:} Our choice of the `Toxic Conversations Set Fit` dataset, originating from Civil Comments, may introduce domain specific biases that do not generalize to the linguistic patterns of other domains where RAG systems are deployed.

\textbf{Evaluation Procedure Consistency:} Potential human bias from manual sentiment verification was mitigated with systematic regex extraction and clear equivalence rules (treating ``mixed'' and ``neutral'' as equivalent). Additionally, our 10\% sampling strategy, necessary for computational constraints, might introduce selection effects.

\textbf{Hyperparameter Selection:} Our chosen generation parameters (\texttt{temperature = 0.1}, \texttt{top\_p = 0.9}) were selected to balance reproducibility and diversity, but we acknowledge that different settings could yield different fairness profiles and alter the magnitude of the observed effects.

\subsubsection{External Validity}

\textbf{Model Generalizability:} While our evaluation covers three diverse SLMs (from Meta, Mistral AI, and NVIDIA), the findings may not generalize to larger (>10B parameters), proprietary (e.g., GPT-4, Claude), or differently trained models.

\textbf{Embedding Model Specificity:} Our use of `all-MiniLM-L6-v2` limits generalizability, as other embedding models or similarity metrics might yield different results. We selected it as it represents a common choice for resource constrained deployments.

\textbf{Language and Cultural Context:} The study is limited to English language content and demographic categories relevant to North American contexts; therefore, findings may not generalize across other languages and cultures.

\textbf{Task Domain Limitations:} Our focus on sentiment analysis provides clear evaluation criteria but may not generalize to other domains (e.g., question answering, code generation) where bias can manifest differently. Our findings serve as a foundation for such investigations.

\textbf{Deployment Environment:} Our controlled experimental setup does not reflect real world conditions like dynamic content or user interaction patterns. Nevertheless, our methodology provides a systematic framework adaptable to production environments.

\subsubsection{Conclusion Validity}

\textbf{Statistical Power and Sample Size:} Our evaluation of 2,100 query pairs per model provides substantial statistical power for detecting fairness violations, with effect sizes (ASR ranging from 17.95\% to 33.00\%) well above typical significance thresholds. The large sample size enables reliable detection of demographic hierarchy patterns and model specific differences.

\textbf{Multiple Comparisons:} Our analysis involves multiple statistical comparisons across models, perturbation types, and evaluation metrics. While we report descriptive statistics rather than formal hypothesis tests, the consistent patterns observed across multiple dimensions (retrieval ASR, generation ASR, demographic hierarchies) strengthen confidence in our conclusions.

\textbf{Reproducibility and Replication:} We provide detailed hyperparameter specifications, exact model versions, and complete experimental procedures to enable replication. Our replication package includes all data processing scripts and evaluation code. However, the stochastic nature of language models means that exact numerical reproduction may require identical hardware and software environments.

\textbf{Measurement Reliability:} The Retriever Robustness Score (RRS) provides consistent, reproducible measurements of retrieval instability, with empirically derived thresholds validated through manual inspection. The metric's mathematical foundation (Euclidean distance on normalized embeddings plus Hamming distance) ensures measurement stability across experimental runs.

\textbf{Construct-Method Alignment:} Our metamorphic testing approach aligns well with our fairness research question, providing direct evidence for demographic sensitivity without requiring extensive labeled datasets or domain expertise. The method's simplicity enhances reproducibility while maintaining evaluation rigor.

\section{Conclusion and Future Work}

Our work advances the fairness testing landscape by targeting a critical intersection: metamorphic testing in Retrieval-Augmented Generation (RAG) pipelines. While prior studies surveyed by Chen et al.~\cite{chen2024fairness} focus on model-level fairness using causal or statistical definitions, we operationalize individual fairness through the Set Equivalence metamorphic relation applied to sentiment classification tasks with demographic perturbations. From an SE perspective, this study contributes to the growing body of knowledge on testing methodologies for AI-enabled systems, demonstrating how traditional software testing techniques like metamorphic testing can be adapted and extended to address emerging challenges in AI system validation.

Distinctively, we introduce the \textit{Retriever Robustness Score (RRS)} (Table~\ref{tab:robustness-score-interpretation}) to isolate fairness violations at the retrieval stage, positioning the retriever as a first-class test artifact. This contrasts with prevailing approaches that overlook non-parametric components in hybrid SLM architectures and aligns with SE principles that emphasize comprehensive component testing and system integration validation. Our findings show that bias propagation begins before generation, reinforcing the need to extend fairness testing workflows to intermediate components, an insight that has broad implications for SE practices in AI system development.

The practical implications of our work extend beyond academic contributions to real-world SE practices. Our study provides actionable guidance for software development teams working with RAG-based systems, particularly in resource-constrained environments where SLMs are increasingly adopted. The systematic methodology we present can be integrated into existing software development lifecycles, enabling continuous fairness monitoring and regression testing for AI-enabled applications. The RRS offers a concrete metric that software quality assurance teams can incorporate into their testing suites, alongside traditional performance and reliability measures.

Furthermore, our findings have significant implications for software architecture and design decisions. The discovery that fairness violations can emerge from retrieval components suggests that software architects must consider fairness as a cross-cutting concern that affects multiple system layers, similar to how security or performance considerations influence architectural choices. This perspective encourages the development of fairness-aware software architectures where bias mitigation strategies are built into the system design rather than applied as post hoc solutions.

Future work should extend this framework to other SE contexts such as automated code generation tools, intelligent documentation systems, and AI-powered development environments where fairness is critical yet underexplored. Research should integrate the approach into established software testing frameworks and CI/CD pipelines, making fairness checks a routine part of SE workflows rather than a specialized research activity. To increase generalization, we will broaden evaluation beyond the two English datasets by exploring additional NLP tasks and languages, complement individual fairness checks with group-based metrics, and assess the engineering effort required for industrial adoption. Overall, our work shifts the lens of fairness testing from the model to the system, offering methodological and practical contributions to fairness assurance in hybrid AI systems an underexplored yet critical direction for safe and equitable deployment.
\section*{Artifact Availability}
\label{sec:rep-pack}

All the necessary data to replicate this work is available in this repository\footnote{\url{https://anonymous.4open.science/r/Metamorphic-SLM-9955/README.md}}. There is also a README.md file with instructions.

\bibliography{cite}

\end{document}